\documentclass[twoside,11pt]{article}
\usepackage{jmlr}

\usepackage{graphicx}
\usepackage{amsmath}
\usepackage{amssymb}
\usepackage{booktabs}
\usepackage{appendix}
\newcommand{\Lagr}{\mathcal{L}}
\usepackage{caption}
\usepackage{subcaption}
\usepackage{multirow}

\usepackage{times}
\usepackage{helvet}
\usepackage{courier}
\title{ARTEMIS: Using GANs with Multiple Discriminators to Generate Art}
\author{James Baker\\
}
\graphicspath{{art/}{figures/}}

\ShortHeadings{ARTEMIS: Using GANs with Multiple Discriminators to Generate Art}{Baker}
\firstpageno{1}

\begin{document} 

\maketitle

\begin{abstract}
\begin{quote}
   We propose a novel method for generating  abstract art. First an autoencoder is trained to encode and decode the style representations of images, which are extracted from source images with a pretrained VGG network. Then, the decoder component of the autoencoder is extracted and used as a generator in a GAN. The generator works with an ensemble of discriminators. Each discriminator takes different style representations of the same images, and the generator is trained to create images that create convincing style representations in order to deceive all of the generators. The generator is also trained to maximize a diversity term. The resulting images had a surreal, geometric quality. We call our approach ARTEMIS (\textbf{ART}istic \textbf{E}ncoder- \textbf{M}ulti-Discriminators \textbf{I}ncluding \textbf{S}elf-Attention), as it uses the self-attention layers and  an encoder-decoder architecture.
\end{quote}
\end{abstract}

\section{Introduction}
\label{sec:intro}

The basic structure of a Generative Adversarial Network (GAN) is to train a generator to produce a realistic \(x' \in D\), for some dimensionality \(D\) from a random distribution \(z\), and a discriminator to label inputs \(x \in D\) as real or false. Typically, the dimensionality \(D\) is usually the dimensions of an image \(L \times W \times 3\), where L and W are integers. However, in the style transfer and texture generation literature, an image can also be represented by its style representation, also known as a feature map. This opens the possibility of using multiple discriminators for one image, with each discriminator using a different shaped feature map extracted from the image. We used a pretrained VGG-19 model for mapping images to their feature maps, thus making this an example of transfer learning. We also implemented visual attention, so the generator could use nonlocal features.

\section{Related Work}
\label{sec:related}

\subsection{Style Transfer}
\label{sec:styletransfer}
Approaches that didn't use neural networks, like \cite{HertzmannStrokes1998}, which could redraw the edges in images to look painted, or \cite{GoochFacial2004}, which could make human faces appear more cartoon-like, were usually limited in the range of styles they could apply to images. The first style transfer algorithm using deep learning was first implemented by \cite{GatysEB15a}. In this paper, style and content representations were extracted from images using the output of intermediate layers of a pretrained VGG-19 network. An image was optimized to minimize the distance between the gramm matrices of the generated image and the gramm matrices of the style and content representations. By adding a photorealistic regularization term to the objective function when optimizing the image, style could be transferred to a content image without losing the realism of the \cite{LuanPSB17}. While a pretrained VGG-19 model is standard for extracting feature representations of images, work has been made to improve the utility of the lighter ResNet for extracting feature representations as well \cite{wangrobustness2021}.

Instead of optimizing an image, GAN approaches like \cite{CycleZhuPIE2017} used a CycleGAN to generate stylized images, using content images as the source to the generator, expanded upon by \cite{LiuArtsy2018}. Style transfer with GANs have also implemented attentional \cite{Park2019ArbitraryST} and residual \cite{XuResGAN2021} components.

\subsection{GANs for Image Generation}
\label{sec:ganimage}
Originally introduced by \cite{goodfellow2014generative}, Generative Adversarial Networks (GANS), and an expansion on them, the Deep Convolutional GAN (DCGAN) \cite{radford2015unsupervised} have been used for writing text \cite{SaeedPoems2019}, composing melodies \cite{KolokolovaIrish2020}, making 3-D models \cite{WuShape3Dim2018} and super-resolution \cite{wangsuperresolution2018}. \cite{ElgammalLEM17}  optimized the generator to not only create convincing samples, but also to make it hard to classify generated samples to belong to a particular artistic category like baroque or impressionism. Works like \cite{TextureNetworksUlyanovLVL16, XianSLFYH17} conditioned their GANs on sketches to generate realistic images. 

\subsection{Texture Generation}
\label{sec:imagesynthesis}

A texture is the " the perceived surface quality of a work of art" \cite{wiki:Texture_(visual_arts)}. Alternatively called style, it refers to things like the shape of lines, thickness of strokes and visual features other than the raw content of an image. Texture Generation is the process of generating new images that have unique textures, distinct from texture synthesis, which is "from an input sample, new texture images of arbitrary size and which are perceptually equivalent to the sample" \cite{RaadDDM17}. In addition to many traditional ways to generate unique textures \cite{DongTextures2020}, machine learning can be used to generate new textures. Works like \cite{LiWandMarkov16}, \cite{PerceptualJohnsonAL16} and \cite{TextureNetworksUlyanovLVL16} used convolutional networks trained to minimize higher-level stylistic loss to generate novel textures. This method was advanced by \cite{StableHistogramsWilmotRB17}
via training the network to minimize histogram loss. While most work has been for flat images, \cite{Deep3DGutierrez2020} expanded texture synthesis to 3-dimensional objects, mapping each point in 3-dimensional space to a color. 

\subsection{Transfer Learning}
\label{sec:transferlearning}
Network-based transfer learning uses a model, or components of a model trained for one task to perform another task, for a new task \cite{Caruana1994,SurveyTan2018}. 
\cite{Wen2019ANT} used a pretrained VGG-19 \cite{Simonyan15} network to extract features and use them to classify mechanical data. Using a pretrained InceptionV3 network, \cite{maskchowdary2020} determined whether a human was wearing a mask or not. \cite{WeedsEspejo2020} compared VGG and Inception for classifying weeds. The pretrained model does not need to be trained on a different dataset than the final model. For example, GANs can be improved by using pretrained layers of an autoencoder \cite{Nagpal2020AGL} or variational autoencoder \cite{HamUnbalanced202} as part of the generator.

\subsection{Visual Attention}
\label{sec:visualattention}
When humans process input from a sequence or an image, we contextualize each part of the input using other parts of the input. However, we do not pay equal attention to every other part of the input \cite{MnihHGK14}. For a textual example, in the sentence ''Alan said he was hungry'', the word ''alan'' is more useful in determining the meaning of ''he'' than the word ''hungry''. For a visual example, in order to guess the location of the right eye on someones face, the most important information would be the location of the left eye, not the length of their beard. Attention, also known as Non-Locality was originally proposed for text \cite{bahdanau2014neural,LuongPM15}, it was then applied to vision \cite{wangnonlocal2017,VaswaniSPUJGKP17}, finding applications in medicine \cite{GuanRadilogist2018}, super-resolution \cite{DaiSecond2019, zhang2018super2018}, and image generation \cite{GregorDGW15,saganzhang2019}.

\subsection{Wasserstein GANs}
\label{sec:wasserstein}
In order to prevent mode collapse, \cite{arjovsky2017} proposed the wasserstein GAN. As an alternative to gradient clipping, \cite{GulrajaniAADC17} introduced a gradient penalty term

\section{Data}
\label{sec:data}

\subsection{Preprocessing}
\label{sec:preprocessing}
The images used were the 80,000+ images in the WikiArt Images dataset \cite{wikiartimagesSalehE15}. Each image was cropped to be \((256 \times 256 \times 3)\). Images that were smaller than \((256 \times 256 \times 3)\) were concatenated with themselves and then cropped. 
Images were preprocessed by using the output of intermediate layers of a VGG-19 network pretrained on the ImageNet dataset. Only the 16 convolutional layers were used. These image style representations were usually different shapes than the original images. For example, a \((256 \times 256 \times 3)\) image would create a \((32 \times 32 \times 512)\) style representation output from the "block4\_conv1" layer.

\section{Network}
\label{sec:network}

\subsection{VGG Layers}
\label{sec:vgg}
Given the pretrained VGG-19 model, \(VGG\), we define a set of its intermediate blocks \(B\), and a set of corresponding dimensions \(DIMS\) corresponding to the output of the corresponding block when the VGG network processes an image   \( \in \mathbb{R}^{256 \times 256 \times 3}\) image. For example \(B=(block1\_conv1,block3\_conv1,block5\_conv1)\), and \(DIMS_B=	(\mathbb{R}^{256 \times 256 \times 64}, 	\mathbb{R}^{64 \times 64 \times 256}, \mathbb{R}^{16 \times 16 \times 12})\). We then use the notation \(VGG(B)\) to mean the style representation outputs of each layer in \(B\) when the VGG processes an image, and therefore

\[VGG_B: \mathbb{R}^{256 \times 256 \times 3} \longrightarrow DIMS_B\]

For example:

\[VGG_{(block1\_conv1,block5\_conv1)}: \mathbb{R}^{256 \times 256 \times 3} \longrightarrow\]
\[(\mathbb{R}^{256 \times 256 \times 64}, 	 \mathbb{R}^{16 \times 16 \times 12})\]

When \(B\) is a singleton set, consisting of a single block, we may also write \(VGG_{block name}\) or \(DIMS_{block name}\), for example:
\[VGG_{block3\_conv1}: \mathbb{R}^{256 \times 256 \times 3} \longrightarrow DIMS_{block3\_conv1}\]

\begin{table}[h!]
\centering
\begin{tabular}{|c c|} 
 \hline
VGG Block & \(DIM_{block}\) \\ 
 \hline\hline
 no block & \(\mathbb{R}^{256 \times 256 \times 3}\)\\ 
 block1\_conv1 & \(\mathbb{R}^{256 \times 256 \times 256}\)  \\
 block2\_conv1 & \(\mathbb{R}^{128 \times 128 \times 128}\)  \\
 block3\_conv1 & \(\mathbb{R}^{64 \times 64 \times 256}\) \\
 block4\_conv1 & \(\mathbb{R}^{32 \times 32 \times 512}\)  \\
  block5\_conv1 & \(\mathbb{R}^{16 \times 16 \times 512}\)  \\
 \hline
\end{tabular}
\caption{VGG Layers and Dimensions}
\label{table:1}
\end{table}

\subsection{Autoencoder}
\label{sec:autoencoder}

The Autoencoder consisted of three parts. First, the encoder \(E\) mapped an image or an image style representation to a latent space representation \(z \in \mathbb{R}^{2 \times 2 \times 64}\). Given the shape of an image or image representation \(DIM_{block}\), where block is the block of the VGG-19 network that produced the style representation, (for example, the shape of the image style representation from \(block1\_conv1\) of the VGG-19 \(\in \mathbb{R}^{256 \times 256 \times 64}\)), we can write:
\[Encoder_{block}: DIM_{block} \longrightarrow \mathbb{R}^{2 \times 2 \times 64}\]

The second part was a decoder that mapped the latent noise to an image.

\[Decoder: \mathbb{R}^{2 \times 2 \times 64} \longrightarrow \mathbb{R}^{256 \times 256 \times 3}\]

The final part was the pretrained \(VGG_{block}\).

The Autoencoder was defined as:

\[Autoencoder_{block}=VGG_{block}(Decoder(Encoder_{block}))\]
\[Autoencoder_{block}: DIM_{block} \longrightarrow DIM_{block} \]

The Encoder used convolutional layers to downsample, and the decoder used Convolutional Transpose blocks to upsample.  After each Convolutional and Convolutional Transpose Block, we used Batch Normalization, to add noise \cite{batchnormIoffeS15} and stabilize the gradients \cite{batchnormsanturkar2019does}, followed by LeakyReLu. While we concede that the swish \cite{switch2017} and related mish \cite{MishMisra2019} activation functions have performed better in some tasks, LeakyReLu was used for the most part due to it being less computationally expensive than the alternatives.

\subsection{Adversarial Components}
To build the generator, we extracted the decoder after pretraining the autoencoder, and attached a \(VGG_B\) instance, where B was the set of blocks we wanted to study. For training, we would generate random noise in \(\in \mathbb{R}^{2 \times 2 \times 64} \) and use that as input to the generator to generate samples.

\[Generator_B=VGG_B(Decoder)\]
\[Generator_B: \mathbb{R}^{2 \times 2 \times 64} \longrightarrow DIMS_B\]

For each \(block \in B\), we had a separate discriminator, that would take an image or style representation and output a probability between 0 and 1

\[Discriminator_{block}: DIM_{block} \longrightarrow [0,1]\]

\section{Experiments}
\label{sec:experiments}

\subsection{Loss Function}
\label{sec:loss}

The autoencoder was trained to minimize reconstruction loss between an image (or style representation) \(x\) and the reconstructed image (or style representations) \(x'\)
\[\Lagr_{AE}=\sum_i^N ||x_i-x_i'||_F\]



The diversity loss penalized the generator for outputting samples that were too similar \cite{Li2017DiversifiedTS}. Given generated samples \(G(z_i),G(z_j),,,G(z_k)\), diversity loss was defined 

\[\Lagr_{DV}= -\sum_i^k  \frac{||G(z_i)-G(z_j)||_F}{||z_i-z_j||_F}, i \ne j\]

In experiments, for each training step, we used k=3 generated samples.

In traditional GAN fashion \cite{goodfellow2014generative}, for each sample, real, \(x_i\) or generated \(G(z_i)\), the Discriminator output a value from 0 to 1, \(D_A\), corresponding to its confidence that the sample was real. 
The Discriminator Adversarial loss was defined, for each \(Discriminator_{block}\) for \(block \in B\)

\[\Lagr_{D_{block}}=\sum_i^N log(D_{block}(x_i))+(1-log(D_{block}(G(z_i)))\]

The generator was trained to fool each Discriminator, so its loss function was defined
\[\Lagr_{GA}=- \sum_{block}^B \sum_i^N log(D_{block}(G(z_i))\]

The total Generator loss to be minimized was thus

\[\Lagr_G=\Lagr_{CE}+\Lagr_{DV}+\Lagr_{GA}\]

\subsection{Training}
\label{sec:training}

The autoencoders were trained for 250 epochs. The Adversarial loop was trained for 300 epochs. The batch size was 8. We added gaussian noise \(\sim \mathcal{N}(0,0.01)\) to labels for binary classification in order to implement label smoothing \cite{salimangoodfellow2016,perturbations}, to reduce the vulnerability of the network to adversarial samples.

\subsection{Subjective Results}
\label{sec:results}
\begin{figure}
    \centering
    \begin{tabular}{c c}
           \includegraphics[scale=.25]{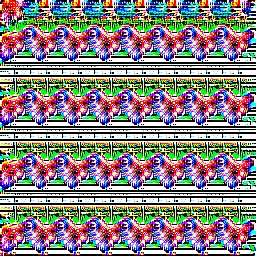} &  \includegraphics[scale=.25]{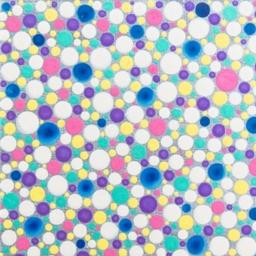}\\
        Image 1 &  Image 2\\
        \includegraphics[scale=.25]{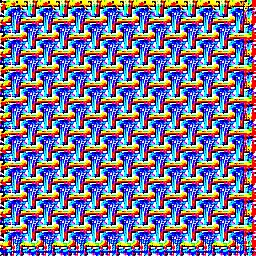} & \includegraphics[scale=.25]{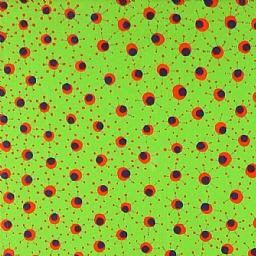} \\
        Image 3 & Image 4\\
        \includegraphics[scale=.25]{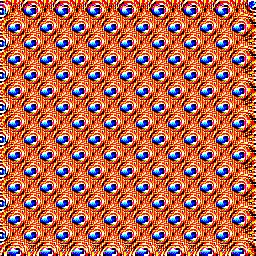} & \includegraphics[scale=0.25]{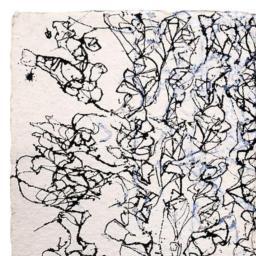} \\
        Image 5 & Image 6\\
        \includegraphics[scale=0.25]{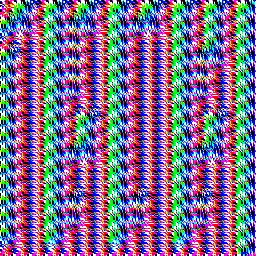} & \includegraphics[scale=0.25]{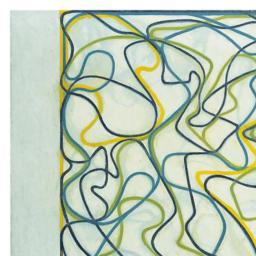} \\
        Image 7 & Image 8 \\
        \includegraphics[scale=0.25]{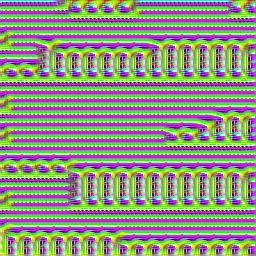} & \includegraphics[scale=0.25]{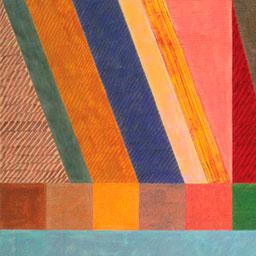}\\
        Image 9 & Image 10 \\
    \end{tabular}
    \caption{The Images used for Comparison}
    \label{fig:images}
\end{figure}

Beauty is in the eye of the beholder, so we decided to evaluate the generated art by comparing it to actual abstract art in the WikiArt dataset. Figure \ref{fig:images} shows the 5 artificial images, all indexed with odd numbers and in the left column, and the 5 manmade images, all indexed with even numbers, in the right column.

An online survey was given out, where 54 participants were asked to rate each image on a scale of 1-10, 10 being the most positive and 1 being the most negative. For each image the average rating and standard deviation are given in table \ref{tab:ratings}. Notably, while the artificial images had much lower ratings, there was less variance in their ratings, suggesting the value of the ARTEMIS model is to produce images that humans will have a predictable reaction to.

\begin{table}[h]
    \centering
    \begin{tabular}{c|c|c}
    Image & \(\mu\) & \(\sigma\) \\
    \hline
        Image 1 & 3.37 & 2.18 \\
        Image 2 & 5.03 & 2.63 \\
        Image 3 & 3.64 & 2.36 \\
        Image 4 & 3.94 & 2.32 \\
        Image 5 & 3.37 & 2.19 \\
        Image 6 & 6.26 & 2.16 \\
        Image 7 & 3.45 & 2.05 \\
        Image 8 & 6.59 & 2.53 \\
        Image 9 & 3.72 & 2.31 \\
        Image 10 & 6.93 & 2.25 \\
    \end{tabular}
    \caption{Caption}
    \label{tab:ratings}
\end{table}

\section{Conclusion}
\label{sec:conclusion}
\subsection{Contributions}\label{sec:contributions}

This paper proposed a novel method of using generative adversarial networks with an ensemble of discriminators, each trained to classify images extracted from a different layer of the pretrained VGG-19 network. Trained on the WikiArt dataset, the network generated images that were akin to abstract art. Compared to manmade art, the artificially generated art was not as well received but there was less variability in how people reacted to the art. 

\subsection{Further Work}\label{sec:furtherwork}
An obvious first step would be to use a new dataset. The WikiArt dataset contains almost no photographs, so it would be worthwhile to see how this model performs when trained using actual images. The WikiArt dataset is also extremely varied in the subjects it portrays and the scale and size involved, so a more homogenous dataset may be used for different results. Applying the multiple discriminator architecture to text would require a different pretrained style extractor than VGG-19, but would be promising.

\bibliography{iccc}
\newpage
\appendix

\section{Architecture}\label{appendix:architecture}

\subsection{Encoder}
Though there were a few different input shapes, we made the encoder map them all to a standard noise size \(\mathbb{R}^{2 \times 2 \times 64}\). The encoder was made of encoder blocks, parameterized by \(c\), an integer representing the number of output channels in the encoder block, and \(s\), the size of the stride:

\begin{figure}[h]
    \centering
    \includegraphics[scale=.25]{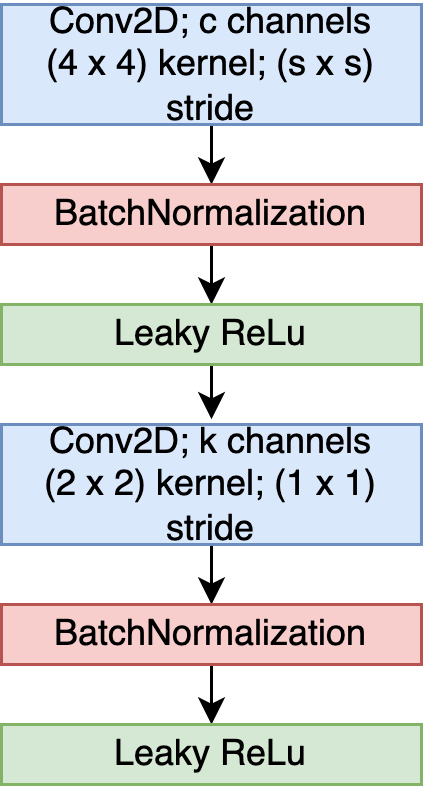}
    \caption{Encoder Block}
    \label{fig:encoderblock}
\end{figure}

First we wanted to make sure every layer had 64 channels. For input dim \(\mathbb{R}^{256 \times 256 \times 3}\), we started with a convolutional block with 8 channels and kernel size \((8 \times 8\) with stride \((1 \times 1\), followed by BatchNormalization and Leaky ReLu. Then we had 3 encoder blocks, each with \(s=1\), with \(c=16,32,64\). For all the other input dimensions (see table \ref{table:1}), we held \(s=1\) constant, and added layers where \(c= \frac{1}{2}\) of the output channels of the last layer, until the output had 64 channels.

Then we wanted to shrink the spatial dimensions to be \((2 \times 2)\). Holding \(c\) constant, we used \(s=2\), (this would halve the spatial dimensions each time), and applied encoder blocks until the output was \((2 \times 2 \times 64)\). We then added normal noise \(\sim \mathcal{N}(0,1)\). Adding this noise will make the decoder more robust and lessen the chance of overfitting.

\subsection{Decoder}

\begin{figure}[h]
    \centering
    \includegraphics[scale=.25]{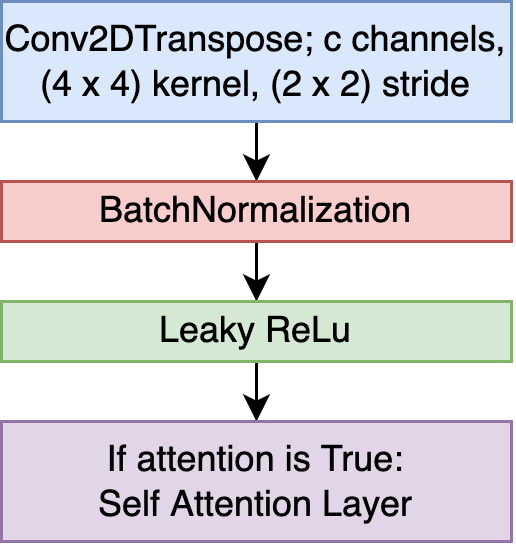}
    \caption{Batch Decoder Block}
    \label{fig:batchdecoderblock}
\end{figure}

The decoder mapped the noise to an image \(\mathbb{R}^{2 \times 2 \times 64} \rightarrow \mathbb{R}^{256 x 256 x 3}\). It was mainly composed of batch decoder blocks shown in figure \ref{fig:batchdecoderblock}, parameterized by the boolean \(attention \in \text{True} || \text{False}\), and \(c\), the number of output channels.

\begin{figure}[h]
    \centering
    \includegraphics[scale=.25]{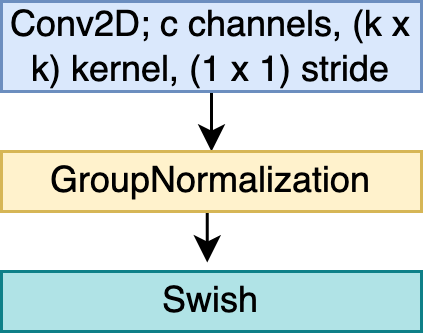}
    \caption{Group Decoder Block}
    \label{fig:groupdecoderblock}
\end{figure}

The decoder also contained group decoder blocks, show in figure \ref{fig:groupdecoderblock}, parameterized by \(c\), the number of output channels, and \(k\), the size of the kernel.

The decoder consisted of 7 batch decoder blocks, where for each block, \(c= \text{max}(32, c')\), where \(c'\) was the amount of output channels in the prior batch decoder block, or the amount of channels in the input in the case of the first batch decoder block, and \(attention\) was True for the first 4 blocks, and False for the last 3 batch decoder blocks. Following that, there were 3 group decoder blocks, with \(c=(16,8,4\) and \(k=(8,8,4)\), respectively. Finally, there was a Convolutional Layer with 3 output channels, a sigmoid activation function, and a rescaling layer to bound the values between 0 and 255, thus producing an image.

\subsection{Discriminator}

\begin{figure}[h]
    \centering
    \includegraphics[scale=0.25]{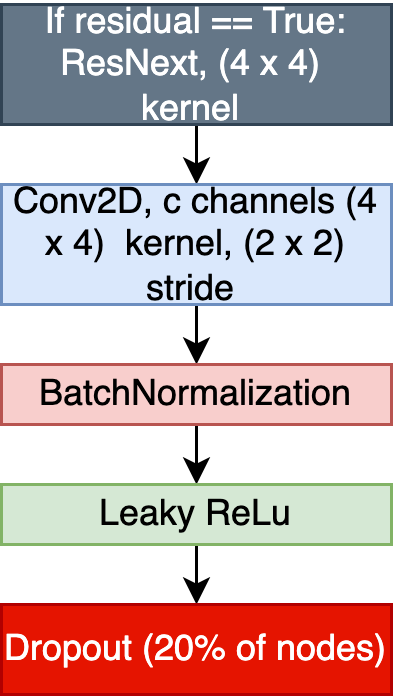}
    \caption{Discriminator Block}
    \label{fig:discblock}
\end{figure}

The discriminator(s) were composed of discriminator blocks shown in figure \ref{fig:discblock}, parameterized by \(c\), the number of output channels and \(resnext\), a boolean of whether to use a ResNext layer \cite{resnextXieGDTH16}.

First, the discriminator had 4 discriminator blocks, each with \(c=\frac{1}{2}c'\), where \(c'\) was the amount of channels in the prior block, or the amount of channels in the input in the case of the first discriminator block, and for all but the first discriminator block, \(resnext\) was True. After that, we flatten the input. Then we apply a fully-connected layer with 8 nodes, batch normalization, leaky ReLu, then the final dense layer with 1 node and sigmoid activation; this final node serves to output the (0,1) label.

\end{document}